\documentclass[fleqn,10pt]{wlpeerj}

\usepackage{multirow} % 표 여러 줄 
\usepackage{makecell} % 표 여러 줄
\usepackage{kotex}    % For Korean
\usepackage{url}
\usepackage{float}    % 표를 본문 바로 다음에

\title{Evaluating Large language models on Understanding Korean indirect Speech acts}

\author[1 2]{Youngeun Koo}
\author[2 3]{Jiwoo Lee}
\author[2]{Dojun Park}
\author[4]{Seohyun Park}
\author[2 3 4]{Sungeun Lee}
\affil[1]{Department of German Language and Literature, Sungkyunkwan University, Seoul, South Korea}
\affil[2]{Artificial Intelligence Institute of Seoul National University(AIIS), Seoul, South Korea}
\affil[3]{Department of German Language and Literature, Seoul National University, Seoul, South Korea}
\affil[4]{Brain Humanities Lab, Seoul National University, Seoul, South Korea}

\corrauthor[2 3 4]{Sungeun Lee}{cristlo5@snu.ac.kr}

\begin{abstract}
To accurately understand the intention of an utterance is crucial in conversational communication. As conversational artificial intelligence models are rapidly being developed and applied in various fields, it is important to evaluate the LLMs’ capabilities of understanding the intentions of user’s utterance. Speech act is a linguistic concept in pragmatics, the study of human language use, and can be simply understood as referring to the intention of an utterance. This study evaluates whether current LLMs can understand the intention of an utterance by considering the given conversational context, particularly in cases where the actual intention differs from the surface-leveled, literal intention of the sentence, i.e. indirect speech acts.
With a specific focus on Korean, a context-sensitive language, we construct evaluation datasets, consisting of three types of indirect speech acts based on Searle’s speech act scheme. The dataset consists with two scenarios, where the same utterance conveys direct speech act and indirect speech act respectively, depending on the conversational context. We adopt two experimental setups: the conventional Multiple-Choice Question(MCQ) format for automatic evaluation, and the Open-Ended Question(OEQ) for detailed assessment by human experts. For thorough evaluation, we also conduct MCQ based experiment on humans to set gold standard. 
Our findings reveal that Claude3-Opus outperformed the other competing models, with 71.94\% in MCQ and 65\% in OEQ, showing a clear advantage. In general, proprietary models exhibited relatively higher performance compared to open-source models. Nevertheless, no LLMs reached the level of human performance. Most LLMs, except for Claude3-Opus, demonstrated significantly lower performance in understanding indirect speech acts compared to direct speech acts, where the intention is explicitly revealed through the utterance. This study not only performs an overall pragmatic evaluation of each LLM’s language use through the analysis of OEQ response patterns, but also emphasizes the necessity for further research to improve LLMs’ understanding of indirect speech acts for more natural communication with humans.
\end{abstract}

\begin{document}

\flushbottom
\maketitle
\thispagestyle{empty}

\section*{Introduction}

ChatGPT, released in November 2022, demonstrated remarkable performance not only in traditional Natural Language Processing (NLP) tasks, such as categorization and translation, but also in more complex tasks, including advanced logical problem-solving and creative writing \citep{sudirjo, orru}. In response, numerous corporations have invested in developing their own models, resulting in the emergence of increasingly sophisticated models. With the increasing number of Large Language Models(hereinafter LLMs) being developed, the need for rigorous performance evaluation has become increasingly critical. These evaluations not only ensure alignment with standards deemed acceptable in contemporary fields but also help identify the strengths and weaknesses of each model \citep{chang}.

Currently, benchmarks are widely employed as standard tools for assessing the capabilities of LLMs \citep{fourrier}. These benchmarks facilitate the evaluation process by providing standardized and automated task-specific datasets. However, the prevailing trend in current benchmarking solely focuses on tasks such as knowledge processing, logical reasoning, and computation \citep{clark, hendrycks}. As LLMs become increasingly integrated into the daily lives of the general public, their performance as personal agents gains paramount importance. This trend necessitates the development of models capable of understanding human language and comprehending user intentions to offer customized assistance effectively.

\begin{table}[ht]
\centering
\begin{tabular}{@{}p{4cm} p{10cm}@{}}
\toprule
\textbf{Utterance} & \textbf{Intention Interpretation} \\ \midrule
\multirow{4}{*}{“The elevator is out of order.”}  
 & \textbf{Context 1:} A janitor walking by points where the stairs are with his/her finger. \\  
 & \textbf{Intention 1:} Suggesting to use the stairs instead of trying to use the elevator. \\ \cmidrule(l){2-2}
 & \textbf{Context 2:} A customer in the store talks to the facility manager. \\  
 & \textbf{Intention 2:} Informing the facility manager about the inconvenience and demanding to solve the problem. \\ \bottomrule
\end{tabular}
\caption{\label{tab:context_utt}Intention interpretation depending on the context}
\end{table}

Pragmatics is a branch of linguistics that examines how language is used and understood within specific contexts and situations. For example, consider the utterance, “The elevator is out of order,” as shown in Table \ref{tab:context_utt}. This statement can convey both a literal and an implied meaning depending on the context. Literally, it means that the elevator is currently non-functional. However, in a particular situation, it may carry an implied meaning, such as serving as a suggestion or direction to take an alternative route, like using the stairs. This example underscores the importance of context-dependent interpretation and illustrates how a speaker utilizes language to convey intentions and implications based on the surrounding context. Simultaneously, in face-to-face interactions, the listener interprets the utterance’s implications and the speaker’s underlying intention. This highlights the essential role of pragmatic competence in facilitating effective communication and managing the dynamic interplay between speaker and listener.

Despite the significant necessity of pragmatic competence in LLMs, there are relatively few evaluation studies specifically addressing this aspect. To bridge this gap, we develope a test set tailored to assess the pragmatic capabilities of LLMs, focusing on their ability to comprehend a speaker’s intention conveyed directly or indirectly through an utterance within its contextual setting. The test suite consists of 240 question units written in Korean, designed to evaluate the models’ proficiency in interpreting a speaker’s intentions based on contextual cues. This focus is particularly significant in the Korean language, where shared context and background information play a crucial role in effective communication \citep{hall76, bhagat, servaes}.\footnote{\cite{hall76} proposes a spectrum of context dependency levels which has low- and high-context systems at each end. In low-context communication, the listener knows very little and must be told practically everything, whereas in high-context communication, the listener is already ‘contextualized’ and does not need to be given much background information \citep{hall90}. According to \cite{bhagat}, nations such as Japan, China, Greece and Spain have those of high-context cultures and those Asian countries show high-context driven communication styles \citep{servaes}.}
This evaluation framework is grounded in Austin and Searle’s Speech Act Theory \citep{austin, searle76}, which categorizes speech intentions into five distinct types: representatives, directives, commissives, expressives, and declaratives. And we assessed how well the LLMs comprehend speakers’ indirect intentions, to be specific, directive, commissive, and expressive intentions realized in representative utterances.

Additionally, we conduct an experiment to evaluate human speech act comprehension capabilities, providing reference data for analyzing and comparing the performance of each LLM. This human reference serves as a critical point of comparison to assess the extent to which LLMs align with human-like understanding of speech acts. 

To summarize, the primary contributions of this paper are:

\begin{itemize}
    \item \textbf{Development of a Speech Act Comprehension Test Set}: A specialized test set designed to evaluate LLMs’ ability to understand and interpret three different category of speech acts within contextual frameworks.\footnote{The test set is publicly available on our GitHub repository at https://github.com/annonymous/.}
    \item \textbf{Collection of Human Test Data}: Human performance data collected to serve as a benchmark for comparison against LLMs’ performance, providing a reference point, i.e. Gold Standard, for evaluation.
    \item \textbf{Systematic Evaluation and Analysis of LLMs’ Performances}: A thorough evaluation and analysis of the LLMs’ capabilities, focusing on their pragmatic competence and alignment with human-like understanding.
\end{itemize}

\section*{Related Work}

\subsection*{Speech Act Theory and Indirect Speech Acts}

It is originally considered by positivist philosophers that every sentence we utter in our everyday life describes or constates something, and thus can always be discerned as true or false. \cite{austin} argued against this concept and categorized our language usage into two types. The first type is the ‘constative utterance’, which can be understood as the same as the traditional idea of a statement. On the other hand, in the case of the second type of language usage called the ‘performative utterance’, sentences have no truth-value. Instead, they become performative acts themselves. For example, uttering “I bet 10 dollars on that” can be considered as an act of ‘betting’, and it cannot be determined whether the sentence itself is true or false. In other words, with performative utterances, we don’t describe the acts we do, but perform the very act itself. We can understand this concept of language usage as ‘Speech Acts’.

\cite{searle79} delved further into Speech Act Theory and categorized speech acts into five classes: representatives, directives, commissives, expressives, and declaratives. The main intention of representative speech acts is to convey or describe something that the speaker believes to be true. For directive speech acts, the intention is to make the listener perform(or not perform) a certain action. Commissive speech acts involve the speaker committing to a future action. Expressive speech acts express the speaker’s emotions about a particular situation. Declarative speech acts create a new state or situation.

Based on the context in which an utterance is made, we can interpret a sentence and the intention of the speaker differently. Take a look at the following utterance: “The ice is thin”. Without a context, it can simply be understood as a description of thin ice, hence as an assertive speech act. However, imagine a situation in which several children are attempting to step on a frozen pond. The same utterance could be interpreted as an act of warning, preventing the children from stepping on the thin ice and falling into the cold water, thus as a directive speech act. We can differentiate between (direct) speech acts and indirect speech acts, which are defined by Searle as utterances in which a certain speech act is performed by performing a different class of speech act. In the example, a directive speech act of ‘advisory warning’ (do not step on the ice) is performed indirectly through an assertive speech act of ‘fact presentation’ (the ice is thin).

As we can see, to understand a speaker’s intention from an utterance, it is crucial to interpret a sentence based on the surrounding context. Considering the existing trend of LLM evaluations through the leaderboards \citep{guo}, which focus primarily on performance in tasks assessing knowledge and understanding explicit literal meanings, evaluating the pragmatic competence of LLMs using the framework of Speech Act Theory can be a meaningful in-depth evaluation of the linguistic performance of LLMs.

\subsection*{Evaluation of human understanding ability of indirect speech acts}
%2-4 문단을 통합 줄여서 2 문단 만들기
%oh 2018(children/learner 화용평가) 설명으로 3 문단 만들기
The evaluation tests of human pragmatic abilities are primarily developed and employed to determine how close the pragmatic competence of the test subjects is to that of general individuals. Most of the assessments focus on individuals with communicative challenges \citep{arcara, kim17, seo, jang}, L1 speaking children and L2 learners \citep{oh}.

\cite{arcara} developed a test for the Assessment of Pragmatic Abilities and Cognitive Substrates(APACS), targeting individuals who have acquired communicative deficiencies. The test comprises the following six tasks in Italian language: Interview, Description, Narratives, Figurative Language 1/2, and Humor. The first task includes autobiographical questions. The Description task involves describing elements of ten everyday pictures. The Narratives task includes comprehension questions about the global topic, specific elements and figurative expressions of a certain story. Figurative Language 1 consists of multiple-choice questions about five idioms, metaphors, and proverbs each. Figurative Language 2 involves open-ended questions. The Humor task includes multiple-choice questions where participants select the appropriate ending(funny, straightforward, unrelated ending) for seven brief stories. The consistency and retest reliability of the test were secured and proved by conducting sample tests with 119 healthy participants representing the general population.

\cite{kim17} developed and conducted a test to evaluate the ability of the students with mild intellectual disability to understand speech acts and indirect speech acts. It is composed of multiple-choice questions, in which certain speech acts are presented in a context that includes two or three sentences. There are three answer options: ‘correct interpretation’, ‘wrong, literal interpretation’ and ‘wrong, context-based interpretation’. Eighteen questions are about indirect speech acts, with half being interrogative sentences and the other half declarative sentences. Another nine questions are about general speech acts, five of which are interrogative sentences and four are declarative sentences. The test was shown to be reliable by producing significantly different results between subjects with varying levels of intellectual challenges.

Other cases of evaluating human understanding of speech act include the tests targeted at children with Asperger syndrome \citep{seo}, right-brain impaired patients \citep{jang} and school-aged children \citep{oh}.

In a similar vein, we developed and employed a speech act comprehension test specifically targeted at LLMs. The main purpose of this test is to determine how well the individual LLMs understand speech acts, with the potential to compare their competence to that of typical human individuals in further research involving human test participants.

\subsection*{Evaluation of LLMs’ understanding ability of indirect speech acts}

Recently, many services are being developed using generative artificial intelligence models in various fields. To assess whether these services can successfully communicate with users and provide requested services, studies on LLM’s ability to understand intention of an utterance are increasing in each field \citep{han, wang, loukas, bouzaki} . In particular, there are many studies using generative artificial intelligence models in language education. Among them, \cite{han} analyzed whether ChatGPT can understand the EFL(English as foreign language) learners’ intention of utterances, which learners enter into ChatGPT while writing English essays. They evaluated whether ChatGPT classifies user input well into 13 predefined intent types such as ‘Request for Translation’, ‘Request for Information’, ‘Statement’, and ‘Acknowledgement’. \cite{wang} examined whether ChatGPT understands which of the talk moves, such as ‘asking for more information’ and ‘making a claim’, correspond to utterances that occur during class, and compared the performance with BERT based models.

% 한문단 추가하기 (교육 외 다른 분야의 LLM 모델 대상으로 화행 이해 평가한 연구) 여행, 금융, 의료(영은), 법률(연구찾아보기, 지우)
% Loukas et al.(2023), Making LLMs Worth Every Penny: Resource-Limited Text Classification in Banking   금융

Moreover, whether LLMs generate appropriate sentences based on a correct understanding of speech acts is dealt with great interest nowadays. Many studies focus on instructing LLMs to generate adequate responses for certain objectives, such as data augmentation and automatic annotation, and evaluating their performance \citep{bouzaki, ostyakova, yu}.

\subsection*{Pragmatic Evaluation of LLMs for Korean}

% 도입부 문단 짧게 추가(한국어 특수성 language specificness \cite{eo} \cite{parka} 강조하기)

\cite{eo} GPT-family models are evaluated on their ability to solve riddles in Korean, a task that demands high creativity and an understanding of language-specific nuances. The results demonstrate that while GPT-4 achieves the highest scores, generally, the models struggle with this task, scoring below 10\% in both EM and F1 scores. This highlights their challenges in tasks that require a deep understanding of the subtleties of human language.

% 같은 대화 기제(그라이스의 협력원리)를 기반으로 LLM 평가하고자 한 프레임웍 연구 -> 한 문장...
\cite{nam} tests Kakao Mini, an AI speaker, for its communicative performance in Korean based on real AI-human conversations in a multi-turn dialogue setup. They leverage the framework of Gricean conversational theory for the evaluation. The results show that the maxim of relation was the most frequently flouted by the tested model, indicating significant room for improvement in achieving natural communication capacity. \cite{park1} evaluates various LLMs, including Korean-specific ones, for their pragmatic competence in Korean, based on Gricean conversational theory. The study assesses whether the models can accurately infer meanings implied by the context, similar to human inference. The findings reveal that while GPT-4 is notable, HyperClovaX, a model tailored for Korean, exhibits superior performance on Korean-specific questions. Building on this, \cite{park2} expand the test suite in size and scope to include other flagship models and additional languages, such as English, Chinese, and German.

\section*{Methods}

\subsection*{Dataset for Evaluation}

To evaluate whether LLMs can understand indirect speech acts, we constructed a dataset consisting of 240 items. Based on Searle’s scheme of speech act categories given in Table \ref{tab:searle5}, we have set three types of indirect speech acts as research objects. \textbf{ReDi} refers to the case where a declarative sentence, which is a ‘\underline{Re}presentative speech act’ as on the surface information, performs a `\underline{Di}rective speech act’ as an illocutionary act. Likewise, \textbf{ReCo} and \textbf{ReEx} refers to the case where a declarative sentence, which is a ‘\underline{Re}presentative speech act’ as on the surface information, performs a `\underline{Co}mmissive speech act’ and ‘\underline{Ex}pressive speech act’ respectively as an illocutionary act. Each indirect speech act type contains 80 items.

\begin{table}[ht]
\centering
\begin{tabular}{@{}cc@{}}
\toprule
\textbf{Category} & \textbf{Examples} \\ \midrule
\makecell{Representative \\ (Assertive)} & \begin{tabular}[c]{@{}c@{}}The earth is round.
\\ This jacket looks good on you. \end{tabular} \\
Directive & \begin{tabular}[c]{@{}c@{}}Please pass me the salt.
\\ Stop running around!\end{tabular} \\
Commissive & \begin{tabular}[c]{@{}c@{}}I promise to finish it by today.
\\ I’ll help you with your homework.\end{tabular} \\
Expressive & \begin{tabular}[c]{@{}c@{}}I’m so sorry!
\\ Welcome to my house!\end{tabular} \\
Declarative & \begin{tabular}[c]{@{}c@{}}I pronounce you husband and wife.
\\ You are fired!\end{tabular} \\ \bottomrule
\end{tabular}
\caption{\label{tab:searle5}Speech Act Categories of \cite{searle76}}
\end{table}

\begin{table*}[ht]
\centering
\begin{tabular}{@{} >{\centering\arraybackslash}m{0.5cm} >{\centering\arraybackslash}m{2.5cm} >{\raggedright\arraybackslash}p{10.5cm} @{}} 
\toprule
\textbf{ID} & \textbf{Type} & \multicolumn{1}{c}{\textbf{Example (English Translation)}} \\ 
\midrule
75 & ReDi-Direct & \parbox[t]{10cm}{철수와 영희는 부부이다. 영희는 밖에 나갈 준비를 하고 있고, 준비를 마친 철수에게 영희는 오늘 날씨가 어떤지 물었다. \\  
이에 철수는 다음과 같이 말했다. \textit{‘밖에 날씨가 좋네’}. \\  
(Cheolsu and Younghee are a married couple. Younghee, preparing to go out, asked Cheolsu, who was already ready, about the weather. \\  
Cheolsu said, \textit{‘It’s nice outside’}.)} \\ 
\midrule
76 & ReDi-Indirect & \parbox[t]{10cm}{철수와 영희는 부부이다. 두 사람은 오랜만에 여유로운 주말을 보내게 되었다. 그러나 철수가 계속 쇼파에만 누워있는 것을 본 영희는 \\  
철수에게 다음과 같이 말했다. \textit{‘밖에 날씨가 좋네’}. \\  
(Cheolsu and Younghee are a married couple. They finally had a relaxing weekend after a long time. Noticing Cheolsu lounging on the couch all day, Younghee said, \textit{‘It’s nice outside’}.)} \\ 
\midrule
97 & ReCo-Direct & \parbox[t]{10cm}{영희는 엄마에게 내일 저녁에 무엇을 먹을지 물었다. 이에 엄마는 다음과 같이 말했다. \textit{‘내일 저녁 메뉴는 소고기야’}. \\  
(Younghee asked her mother what they would eat for dinner tomorrow. Her mother said, \textit{‘Tomorrow’s dinner menu is beef steak’}.)} \\ 
\midrule
98 & ReCo-Indirect & \parbox[t]{10cm}{영희는 엄마에게 소고기를 먹고 싶다고 말했다. 이에 엄마는 다음과 같이 말했다. \textit{‘내일 저녁 메뉴는 소고기야’}. \\  
(Younghee told her mother that she wanted to eat beef steak. Her mother said, \textit{‘Tomorrow’s dinner menu is beef steak’}.)} \\ 
\midrule
171 & ReEx-Direct & \parbox[t]{10cm}{영희는 민수에게 철수의 직업이 무엇인지 물었다. 이에 민수는 다음과 같이 대답했다. \textit{‘철수는 요리사야’}. \\  
(Younghee asked Minsu what Cheolsu’s job is. Minsu replied, \textit{‘Cheolsu is a chef’}.)} \\ 
\midrule
172 & ReEx-Indirect & \parbox[t]{10cm}{철수가 집들이를 해서 집들이 음식을 만들었다. 철수가 만든 음식을 먹고, 한 친구가 다음과 같이 말했다. \textit{‘철수는 요리사야’}. \\  
(Cheolsu hosted a housewarming party and cooked food for his friends. After tasting the food Cheolsu had made, one of the friends said, \\  
\textit{‘Cheolsu is a chef’}.)} \\ 
\bottomrule
\end{tabular}
\caption{\label{tab:data-example}Examples of Dataset}
\end{table*}

This study aims to evaluate whether LLMs understand and distinguish that the same utterance can be interpreted with different meanings depending on the context. Therefore, we organized every 80 items for each type of indirect speech act into 40 pairs, where the same utterance performs direct and indirect speech act. Examples of our dataset are given in Table \ref{tab:data-example}. 

As shown in Table \ref{tab:data-example}, each item consists of context and utterance. Through the utterance in item \#75, speaker intends to simply provide information through a declarative statement. In its context where the conversation partner is inquiring about the weather, the utterance “it’s nice outside” performs the speech act of ‘providing information’. 
However, in item \#76, the utterance “it’s nice outside” in the given context does not primarily perform the speech act of ‘providing an information’, which is typically associated with declarative statements. Instead, it implicitly, indirectly performs ‘suggesting’ to go outside. This is an example of \textbf{ReDi} where assertive utterance has been uttered to perform the directive speech act.

The context for item \#97 involves a question about the dinner menu and the utterance presents response to this inquiry, making the item \#97 an example of a direct speech act.
In contrast, in item \#98, the example of \textbf{ReCo}, the mother indirectly performs a ‘promise’ through a declarative utterance, that she will prepare beef steak as a dinner menu for Younghee. Here, the utterance conveys an indirect speech act by performing a commitment through declarative utterance.

Similarly, the context for item \#171 asks about the dinner menu and the utterance responds to this question, making the item \#171 another example of a direct speech act.
Despite that, item \#172 is an example of \textbf{ReEx} because the utterance ‘Cheolsu is a chef’ is intended to praise Cheolsu for cooking like a professional chef, rather than to state that Cheolsu’s job is a chef.

\subsection*{LLMs under Evaluation}

In our study, we compare 12 different LLMs divided into two categories. Table \ref{tab:models} lists eight proprietary models accessed via API and four open-source models, each around the size of ten billion parameters, for which we accessed the model weights directly. Among the proprietary models, we include two versions of GPT \citep{gpt}, three versions of Claude \citep{claude} and three versions of Mistral \footnote{\url{https://mistral.ai/}}. We exclude Gemini by Google from our scope due to limited API access. The open-source models feature Llama3-8B(hereinafter Llama3)\citep{llama}, Qwen1.5-14B(hereinafter Qwen)\citep{qwen}, Solar-10.7B \citep{solar}, and T3Q-ko-solar \footnote{\url{https://huggingface.co/chihoonlee10/T3Q-ko-solar-dpo-v7.0}}. Solar-10.7B(hereinafter Solar) is developed by Upstage, a Korean company, while T3Q-ko-solar represents a fine-tuned version of Solar, which was ranked at the top of the Open Ko-LLM Leaderboard \citep{ko-llm} as of May 17, 2024.

\begin{table}[H]
\centering
\begin{tabular}{@{}lll@{}}
\toprule
\textbf{Type} & \textbf{Model} & \textbf{Version} \\ \midrule
\multirow{8}{*}{Proprietary} 
 & GPT-3.5 & turbo-0125 \\
 & GPT-4 & turbo-2024-04-09 \\
 & Claude3-Haiku & haiku-20240307 \\
 & Claude3-Sonnet & sonnet-20240229 \\
 & Claude3-Opus & opus-20240229 \\
 & Mistral-small & small-2402 \\
 & Mistral-medium & medium-2312 \\
 & Mistral-large & large-2402 \\ \midrule
\multirow{4}{*}{Open-Src.} 
 & Llama3-8B & Instruct \\
 & Qwen1.5-14B & 1.5-14B-Chat \\
 & Solar-10.7B & Instruct-v1.0 \\
 & T3Q-ko-solar & dpo-v7.0 \\ \bottomrule
\end{tabular}
\caption{\label{tab:models}Overview of Proprietary and Open-Source LLMs Evaluated}
\end{table}

\subsection*{Experimental Setup and Evaluation Metrics}

We use two different setup for our experiment to assess LLM’s ability to understand indirect speech acts: multiple-choice questions(MCQs) and open-ended questions(OEQs). In the MCQ experiment type, we attempted to evaluate whether LLM can select among the given options what the speech act of the utterance is, given the context and utterance. Four options are presented for LLMs to select from. One is the correct speech act type, another is the ‘opposite type’, and the others are two randomly selected speech act types. ‘Opposite type’ refers to direct speech act type when the test item is regarding indirect speech act, and vice versa. By having direct speech act as one of the option for indirect speech act context and utterance, and vice versa, we seek to evaluate whether LLMs can distinguish that the utterance can perform either indirect speech act or direct speech act depending on the given context.

For MCQ experiment, the performance of LLMs is calculated by the ‘agreement rate’, whether the option selected by LLMs is same with the correct answer. Considering that the response of the LLMs may randomly change each time, we ran through three trials for each model and test units, for robust and reliable assessment. Therefore, the performance of each model in MCQ setting is the percentage of correct answers among all responses, obtained on average by three trials for each of the 240 test units.

In the OEQ experiment type, the responses generated by each LLM were qualitatively assessed by human evaluators. The LLM’s responses to the OEQ test were scored on a scale of 1 to 5, in the aspect of pragmatic intent and literal meaning of an utterance.
A score of 5 denotes a ‘Correct Answer,’ indicating that the LLM accurately grasped both the intent and meaning of the utterance. A score of 4, labeled as an ‘Acceptable Answer,’ applies when the LLM correctly understood either the intent or the meaning, while the other was interpreted in somewhat awkward manner. A score of 3, or ‘Partial Answer,’ is assigned when the LLM correctly understood either the intent or the meaning, but misinterpreted the other. A score of 2, termed a ‘Incorrect, but Partially Acceptable Answer,’ indicates that the LLM misunderstands one of the aspects, and even shows only an unnatural, awkward understanding of the other. A score of 1 represents an ‘Incorrect Answer,’ assigned when the LLM fails to comprehend both the intent and meaning.
Each test unit was evaluated by three human evaluators, and the final score for OEQ test response was calculated as the average of their scores.\footnote{Intraclass correlation coefficient score rated 0.715, based on ICC3k Average fixed raters metrics.} For ease of comparison with MCQ scores, OEQ scores were rescaled, ranging from 0 to 100.

To establish a gold standard, we collected human reference data by conducting a same MCQ test on humans which was held on LLMs. The participants for the experiment were recruited from the general public, who are 20~30 years old and completed the regular South Korean education curriculum with South Korean nationality. To ensure representativeness, those who majored in linguistics or related disciplines were excluded. A total of 52 participants took part in the experiment, and their detailed demographic information is given in Table \ref{tab:participants2}. According to Table \ref{tab:participants2}, there is no significant difference in the performance score of understanding speech act between subjects based on gender, age, education level, major, etc. Hence, this indicates that the human reference data obtained in this study are sufficiently representative.

\begin{table}[H]
\centering
\begin{tabular}{@{}llcc@{}}
\toprule
\textbf{Group} & \textbf{Attribute} & \textbf{Number of Participants} & \textbf{Performance} \\ \midrule
\multirow{2}{*}{Gender} 
 & Male & 24 & 77.12 \\
 & Female & 28 & 78.08 \\ \midrule
\multirow{2}{*}{Age} 
 & 20~24 & 29 & 78.01 \\
 & 25~30 & 23 & 77.94 \\ \midrule
\multirow{3}{*}{\makecell{Academic \\ Background}}
 & Undergraduate Students & 30 & 78.4 \\
 & Undergraduate Graduates & 21 & 76.39 \\
 & Not Applicable & 1 & 72.92 \\ \midrule
\multirow{8}{*}{Major}
 & Humanities & 12 & 76.25 \\
 & Social Sciences & 9 & 77.41 \\
 & Educations & 2 & 80.63 \\
 & Natural Sciences & 4 & 79.79 \\
 & Engineering & 14 & 76.67 \\
 & Arts & 3 & 79.72 \\
 & Medical Science & 2 & 78.13 \\
 & Others & 6 & 79.38 \\ \bottomrule
\end{tabular}
\caption{\label{tab:participants2}Demographic Information on Experiment Participants}
\end{table}

\section*{Result}

\subsection*{Quantitative Analysis on LLM Performance}

\paragraph{Overall Performance} 

\begin{table}[ht]
\centering
\begin{tabular}{@{}llc@{}}
\toprule
\textbf{Type} & \textbf{Model} & \textbf{Performance} \\ \midrule
\multirow{8}{*}{Proprietary} 
 & GPT-3.5 & 50.14 \\
 & GPT-4 & 71.39 \\
 & Claude3-Haiku & 33.75 \\
 & Claude3-Sonnet & 64.31 \\
 & Claude3-Opus & \textbf{71.94} \\
 & Mistral-small & 65.56 \\
 & Mistral-medium & 66.67 \\
 & Mistral-large & 69.58 \\ \midrule
\multirow{4}{*}{Open-Src.} 
 & Llama-3-8B & 49.44 \\
 & Qwen-14B & 40.14 \\
 & Solar-10.7B & \textbf{65} \\
 & T3Q-ko-solar & 59.86 \\ \midrule
\multirow{1}{*}{Human} 
 &  & \textbf{77.64} \\ \bottomrule
\end{tabular}
\caption{\label{tab:result_mcq}Scores on MCQ Test}
\end{table}

Table \ref{tab:result_mcq} illustrates the performance of each LLM on the MCQ test. In general, proprietary models outperform open source models. Among the proprietary models, Claude3-Opus achieved the highest accuracy with score of 71.94, with GPT-4 and Mistral-large closely following with 71.39 and 69.58, respectably. Flagship models generally outperform lightweight models. This result confirms that models trained with a greater number of parameters excel in speech intention inference task, which requires an understanding of subtle contextual nuances. Claude3-Haiku showed the lowest performance among both the proprietary models and Claude family models. In contrast, another lightweight model, Mistral-small, showed no significant difference compared to other models in family. 

Solar demonstrated the highest performance among open-source models, achieving a score of 65.0. However, compared to the top-performing model, Claude3-Opus, it falls short by 7 points and is comparable to Claude3-Sonnet, Mistral-small, and Mistral-medium. T3Q-ko-solar, which follows Solar, is a model fine-tuned based on Solar, but its score dropped by nearly 5 points. While fine-tuning led to improved performance in the Ko-LLM benchmark tasks \citep{ko-llm}, it was found that performance decreased in context-based inference.

Notably, the human score for speech act comprehension was 77.64, which was clearly higher than all LLMs’ scores. Human participants showed a total score of 77.5, achieving a higher performance than Claude3-Opus, which had a total score of 71.94. This indicates that human’s capability of understanding speech acts still surpasses the highest-performing LLM, Claude3-Opus.

To conduct an in-depth analysis on the models’ ability to understand indirect speech acts, a selected subset of models were re-evaluated using OEQ tasks, which requires to directly generate answers without giving pre-defined options. First, the top-performing models each from proprietary and open source type, Claude and Solar, were selected. Next, one of the low performing model from the proprietary models (GPT-3.5) along with an open model which exhibits similar performance to it (Llama3) were chosen for further evaluation. As previously explained, the LLM responses in the OEQ test were evaluated manually by human evaluators.

\begin{table}[ht]
\centering
\begin{tabular}{@{}llc@{}}
\toprule
\textbf{Type} & \textbf{Model} & \textbf{Performance} \\ \midrule
\multirow{2}{*}{Proprietary} 
 & GPT-3.5 & 78.88 \\
 & Claude3-Opus & 90.37 \\ \midrule
\multirow{2}{*}{Open-Src.} 
 & Llama-3-8B & 75.64 \\
 & Solar-10.7B & 86.24 \\ \bottomrule
\end{tabular}
\caption{\label{tab:result_oeq}Scores on OEQ Test}
\end{table}

\begin{figure}[H]
\centering
\includegraphics[width=13cm]{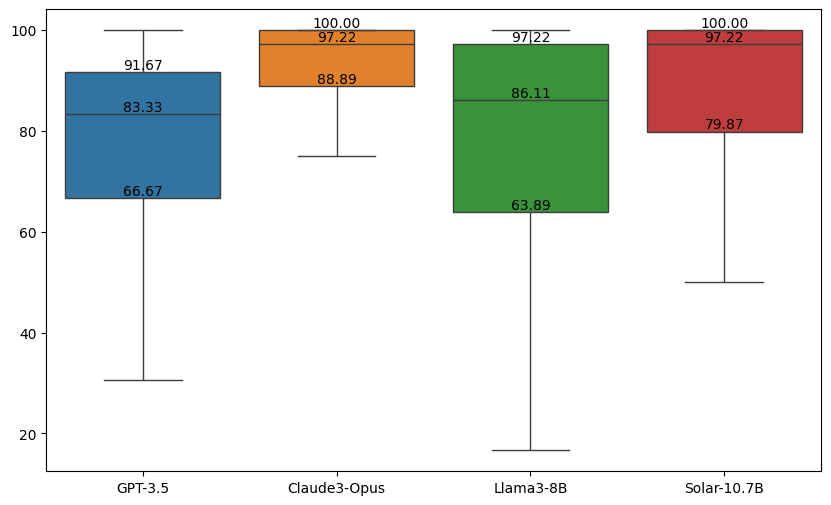}
\caption{OEQ Score Distribution by each LLM}
\label{fig:oeq_distribution}
\end{figure}

Table \ref{tab:result_oeq} presents the OEQ scores by each LLMs. Similar to the MCQ task, Claude3-Opus showed the best performance among the models, Solar following behind. The subtle difference in performance between GPT-3.5 and Llama3 in the MCQ test is also observed in the OEQ test, with only a 3-point difference. Interestingly, the ranking of average accuracy among the models was consistent in OEQ and MCQ result, i.e., in the order of Claude3-Opus, Solar, GPT-3.5, and Llama3. This was likewise observed in terms of the stability of the models. Figure \ref{fig:oeq_distribution} is box plots that illustrate the score distributions of each LLM in the OEQ task. Claude3-Opus showed not only the highest accuracy, as in Table \ref{tab:result_oeq}, but also demonstrates highest stability, given that the Interquartile Range(IQR), the difference between Q1(25\% percentile) and Q3(75\% percentile), is the smallest. Considering the accuracy score and the stability of the model, it is found that the level of speech act comprehension is superior in the order of  Claude3-Opus, Solar, GPT-3.5, and Llama3.

% GPT-3.5 may be better in terms of consistency because the standard deviation of GPT-3.5 is smaller than Solar-10.7B, which indicates that the scores of GPT-3.5 are distributed more evenly. However, Solar-10.7B shows a high score distribution and a low IQR. This means that Solar-10.7B maintains the upper score range stably and is considered to be a more stable model.

\paragraph{Directive vs. Commissive vs. Expressive speech act} 

Scores on MCQ test across three indirect speech act types, \textbf{ReDi}, \textbf{ReCo}, and \textbf{ReEx}, are listed in Table \ref{tab:result_mcq2}. The values in bold represent the maximum values, while the underscored values represent the minimum values. As shown in the Table \ref{tab:result_mcq2}, all models achieved highest accuracy for \textbf{ReEx}, where declarative utterance intends expressive speech act. Most items for \textbf{ReEx} indirect speech act, in our dataset, consist of utterances using idiomatic expressions. For example, the Korean expression ‘꿀맛이야(It’s honey flavored)’ is often used to indicate that something is very delicious. Conversely, ‘개밥이야(It’s a dog food)’ refers that the quality of a certain food is low or unappetizing to eat. In Korea, a dog’s food is often seen as unappealing because it consists of various kibbles mixed together roughly.

Declarative sentences with expressive indirect speech act in our dataset were constructed manually by humans. Therefore, test items include natural expressions of indirect speech acts used by humans, and consequently, score for human on \textbf{ReEx} indirect speech act type was the highest compared to \textbf{ReDi} and \textbf{ReCo} analyzed in Table \ref{tab:result_mcq2}. However, not only humans but all LLMs had the highest score for \textbf{ReEx} among the three types of indirect speech acts. This suggests that LLMs learned a considerable amount of Korean idiomatic expressions and reflect the tendency of human showing the highest scores  for \textbf{ReEx}.

In directive indirect speech sentences(\textbf{ReDi}), which have the intent to demand something, both humans and all LLMs performed the least. To be specific, declarative sentences that describe facts or objective information were difficult to understand as requests, compared to the sentences that describe one’s opinion or subjective information. For example, a staff member could say, “It’s sold out” to customers waiting in line outside a restaurant, to suggest the customers to go to another restaurant. This example is a test sample that all 52 experiment participants and 12 LLMs failed to interpret correctly. All of them understood this utterance as ‘providing information’ rather than ‘suggesting’. However, in the test sample where the speaker pointed to the air conditioner and said, “It’s too cold,” most of the experimental participants (51 out of 52) and LLMs (10 out of 12) analyzed it correctly as ‘requesting’ rather than ‘claiming’. Likewise, when a declarative sentence describes facts rather than asserting opinions, it is challenging to interpret it as a directive speech act. Given that nearly 63\% of the evaluation data in this study consists of \textbf{ReDi} type utterances that depict factual information, both LLMs and human evaluators demonstrated the lowest comprehension in \textbf{ReDi} indirect speech act. 

Another key finding from Table \ref{tab:result_mcq2} is that humans show relatively small deviations across different types of indirect speech acts, whereas the majority of LLMs exhibit larger deviations. Especially lightweight models showed great difference in scores between indirect speech act types.

\begin{table}[ht]
\centering
\begin{tabular}{@{}llccc@{}}
\toprule
\textbf{Type} & \textbf{Model} & \multicolumn{3}{c}{\textbf{Performance}} \\ 
\cmidrule(lr){3-5} &  & \textbf{ReDi} & \textbf{ReCo} & \textbf{ReEx} \\ \midrule
\multirow{8}{*}{Proprietary} 
 & GPT-3.5 & \underline{19.17} & 40.00 & \textbf{50.8} \\
 & GPT-4 & \underline{38.33} & 45.83 & \textbf{57.5} \\ 
 & Claude3-Haiku & \underline{18.33} & 23.33 & \textbf{35.83} \\ 
 & Claude3-Sonnet & \underline{23.33} & 53.33 & \textbf{65} \\ 
 & Claude3-Opus & \underline{63.33} & 73.33 & \textbf{87.50} \\ 
 & Mistral-small & \underline{38.33} & 45.83 & \textbf{57.5} \\ 
 & Mistral-medium & \underline{25} & 47.5 & \textbf{65} \\ 
 & Mistral-large & \underline{37.5} & 54.17 & \textbf{75} \\ \midrule
\multirow{4}{*}{Open-Src.} 
 & Llama-3-8B & 25.83 & \underline{20.83} & \textbf{45} \\ 
 & Qwen-14B & \underline{14.17} & 25 & \textbf{46.83} \\ 
 & Solar-10.7B & \underline{32.5} & 38.33 & \textbf{65.83} \\ 
 & T3Q-ko-solar & \underline{20} & 43.33 & \textbf{59} \\ \midrule
\multirow{1}{*}{Human} 
 &  & \underline{72.45} & 76.92 & \textbf{90.77} \\ \bottomrule
\end{tabular}
\caption{\label{tab:result_mcq2}Scores on MCQ Across Indirect Speech Act Types}
\end{table}

\paragraph{Direct vs. Indirect speech act} 

Table \ref{tab:result_mcq3} describes how well LLMs understand direct and indirect speech acts, respectively. Most LLMs show significantly lower accuracy in indirect speech acts, compared to direct speech acts. While other models exhibited a significant score gap of 31.64\%p, on average, between direct and indirect speech acts, only Claude3-Opus, who ranked first for highest performance, showed balanced accuracy of direct and indirect speech acts. The score difference is only 5.55\%p and Claude3-Opus even showed higher accuracy in indirect speech acts. 

In the case of humans, scores for both direct and indirect speech acts were relatively high and balanced. In particular, the comprehension of indirect speech acts was slightly higher, and therefore, Claude3-Opus was the only model that showed similar tendencies to humans.

\begin{table}[ht]
\centering
\begin{tabular}{@{}lllcc@{}}
\toprule
\textbf{Type} & \textbf{Model} & \multicolumn{2}{c}{\textbf{Performance}} \\ 
\cmidrule(lr){3-4} &  & Direct & Indirect \\ \midrule
\multirow{8}{*}{Proprietary} 
 & GPT-3.5 & \textbf{63.61} & 36.67 \\
 & GPT-4 & \textbf{84.72} & 58.06 \\ 
 & Claude3-Haiku & \textbf{41.94} & 25.83 \\ 
 & Claude3-Sonnet & \textbf{81.39} & 47.22 \\ 
 & Claude3-Opus & 69.17 & \textbf{74.72} \\ 
 & Mistral-small & \textbf{83.89} & 47.22 \\ 
 & Mistral-medium & \textbf{87.50} & 45.83 \\ 
 & Mistral-large & \textbf{83.61} & 55.56 \\ \midrule
\multirow{4}{*}{Open-Src.} 
 & Llama-3-8B & \textbf{68.33} & 30.56 \\ 
 & Qwen-14B & \textbf{51.67} & 28.61 \\ 
 & Solar-10.7B & \textbf{84.44} & 45.56 \\ 
 & T3Q-ko-solar & \textbf{78.89} & 40.83 \\ \midrule
\multirow{1}{*}{Human} 
 &  & 75.22 & \textbf{80.05} \\ \bottomrule
\end{tabular}
\caption{\label{tab:result_mcq3}Scores on MCQ for Direct and Indirect Speech Act}
\end{table}

% OEQ에서 직접, 간접화행 비교는 생략
%
% \begin{table}[H]
% \caption{\textbf{Scores on OEQ Across Indirectness}}
% \centering
% \begin{tabular}{@{}lllcc@{}}
% \toprule
% \textbf{Type} & \textbf{Model} & \multicolumn{2}{c}{\textbf{Performance}} \\ 
% \cmidrule(lr){3-4} &  & Direct & Indirect \\ \midrule
% \multirow{2}{*}{Proprietary} 
%  & GPT-3.5 & \textbf{79.75} & 78 \\ 
%  & Claude3-Opus & 88.58 & \textbf{92.25} \\ \midrule
% \multirow{2}{*}{Open-Src.} 
%  & Llama-3-8B & \textbf{82.33} & 69 \\ 
%  & Solar-10.7B & \textbf{91.83} & 80.58 \\  \bottomrule
% \end{tabular}
% \label{tab:result_oeq3}
% \end{table}

\paragraph{Error Analysis} 

\begin{figure}[ht]
\centering
\includegraphics[width=\linewidth]{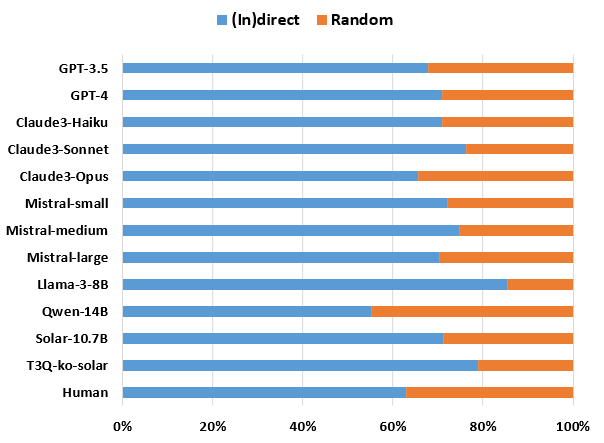}
\caption{MCQ Error Choice Distribution by each LLM}
\label{fig:result_mcq4}
\end{figure}

Figure \ref{fig:result_mcq4} demonstrates the distribution of which incorrect answer options each model selected during MCQ task, informing the tendency of incorrect answers chosen by LLMs. This figure aims to reveal whether LLM has difficulty understanding speech acts in a given context, or whether LLM cannot understand the input itself, language expression. 
Figure \ref{fig:result_mcq4} shows the ratio of two types of errors, ‘(In)direct’ and ‘Random’, among the total errors. This shows which type of error LLM makes more frequently. The MCQ type among the test data constructed in this study consists of four options: one is the type of speech act when an utterance is understood as a direct speech act, another is the type of speech act when an utterance is understood as an indirect speech act, and the rest are two types of random speech acts. ‘(In)direct’ is when an utterance that is actually a direct speech act is misunderstood as an indirect speech act or vice versa, and ‘Random’ is when one of the two random options is misunderstood.

As can be seen in Figure \ref{fig:result_mcq4}, most models made errors by confusing direct and indirect speech acts rather than selecting random options. Except for the Qwen-14.B model, which has a significantly low accuracy and thus does not make much sense to analyze the tendency of error types, we can see that the flagship models, such as Claude3-Opus, Mistral-large, and top open-source model, Solar, made fewer mistakes caused by confusion between direct and indirect. Humans were the least likely to confuse direct and indirect, and predominantly excelled at context-dependent interpretation, compared to LLMs.

\subsection*{Qualitative Analysis on LLM Performance}

The ranking of average accuracy among the models was consistent in OEQ and MCQ result, i.e., in the order of Claude3-Opus, Solar, GPT-3.5, and Llama3. Therefore, we conducted a qualitative analysis on the responses of LLMs in MCQ and OEQ tests.

\paragraph{Explanatory Capabilities of LLMs in MCQ Task} 

% 후보 1: Problem Solving of LLMs in MCQ Task
% 후보 2: Explanatory Proficiency of LLMs in MCQ Task
% 후보 3: Explanatory Depth of LLMs in MCQ Task

In the MCQ test, we evaluated and scored each model’s interpretative capabilities, specifically assessing whether the model selected the correct option number from the answer choices. However, given the generative nature of LLMs, their ability to produce relevant explanations and contextual information should also be considered when assessing overall performance \citep{khatun}.

Open-source models overall adhered closely to prompt instructions, generating responses limited to the answer choice and its corresponding number. In the few instances where they deviated, the responses contained irrelevant contents straying away from the correct answer. Notably, T3Q-ko-solar-10B distinctively demonstrated this tendency, generating not only the option number but also additional comments and explanations — regardless of their correctness — in 94 out of a total of 240 cases.

The other models — Llama3, Solar, and Qwen — formulated their answers in alignment with the prompt instructions, correctly following the directive ‘to choose one of the options.’ Notably, Qwen exhibited a distinctive answer pattern, in some cases providing only the specific number of the selected option, without including the name of each option’s element.

Proprietary models demonstrated distinct tendencies in performance and response patterns associated with their respective developing companies. The three Mistral models, regardless of accuracy rate, strictly adhered to prompt instructions by selecting and generating only the option number and its name. Similarly, GPT-3.5, as a prominent proprietary model alongside GPT-4 and Claude3-Opus, produced answers well-suited to MCQ formats, demonstrating a comparable adherence to prompt instructions, similar to open-source models. In contrast, GPT-4 and Claude models often went beyond simply selecting an option, frequently providing additional explanations to elucidate the inference process for users. This behavior suggests a more user-oriented approach in GPT-4 and Claude, emphasizing transparency in reasoning alongside task performance.

\paragraph{Response Pattern by LLMs in OEQ Task} 

In the open-ended question test set, we evaluated the overall generative capabilities of two proprietary and open-source language models. Through an analysis of the responses, we identified common answer patterns and specific challenges where each model encounters difficulties in interpreting the intended meaning of speech within a given context.

Llama3 exhibited a predominant error tendency by generating random, irrelevant tokens in nearly all of its responses, many of which included unreadable emoticons and emojis. Furthermore, although all prompts in the test set were presented in Korean, the majority of Llama3’s responses — an average of 185 out of 240 (77 percent) — were generated in English. This suggests that the model is possibly overfitted to English prompts, assuming users are native English speakers without knowledge of Korean language. This behavior indicates a potential misalignment with the language context specified in the prompts’ instructions.

Another open-source model, Solar, exhibited a similar error pattern to Llama3, though with fewer occurrences. A unique but non-dominant characteristic of Solar was its tendency to generate its own answer choices, such as (A), (B), and (C), and then select an answer from these, mimicking the response format of an MCQ test. This behavior occurred an average of 17 times across three repeated responses, deviating from the typical approach humans use for similar tasks. This suggests that Solar may be overfitted to MCQ test sets that mirror benchmark formats, likely as a strategy to optimize its performance on leaderboard evaluations.  This answer pattern appears to be closely associated with its inclination to generate one or a few definitive answers. Furthermore, Solar exhibited a tendency to produce Chinese Hanja characters, which are no longer commonly used in contemporary Korean language, reflecting a possible bias or outdated language modeling in its training.

While achieving a score comparable to Solar on the OEQ test set, GPT-3.5’s responses were written in Korean and included significantly fewer random or unreadable outputs than those of Solar. GPT-3.5 consistently provided one or a few decisive answers that demonstrated an accurate understanding of the intended meaning of the speech. These subtle yet notable differences between Solar and GPT-3.5 highlight the importance of evaluating not only the comprehension abilities but also the generative quality of LLMs. Such evaluations provide a more comprehensive and accurate assessment of their overall performance.
Claude3-Opus, which achieved the highest score among the models, demonstrated its advanced capabilities by offering multiple detailed explanations regarding speech intentions in specific contexts, often using a list format. Additionally, it provided supplementary information to enhance the user’s understanding of the answers. This tendency reflects the model’s high-level language capabilities, enabling it to perform sophisticated analyses and provide nuanced interpretations of complex linguistic tasks.

\section*{Discussion}

\subsection*{Indirect Comprehension Capabilities of LLMs}

% - Multiprag 논문이랑 비교 (함축)
% - 오픈(3): llama3, qwen, solar
% - 상업(8): gpt3.5, gpt4, claude3, claude3, claude3, mistral, mistral, mistral

% 대화함축 결과와 비교
\cite{park1, park2} analyzed LLM’s ability to understand conversational implicature. They analyzed how much LLM understands conversational implicature according to Grice’s four conversational maxims: quality, quantity, relation, and manner \citep{grice}.\footnote{Conversational implicature refers to the meaning that is implied in a conversation, which is not directly stated but inferred based on context, shared knowledge, and conversational norms.} Implicature is one of the main research topics of pragmatics along with speech acts, and conversational implicature is closely related to speech acts, which is the subject of this paper, in that it refers to the meaning of a sentence that is not explicitly expressed in an utterance but can be understood from a given conversational context. Accordingly, in order to compare LLM’s speech act understanding and LLM’s conversational implicature understanding, this paper analyzed the correlation between performances, as shown in the Table \ref{tab:discussion_correlation}.

% 상관관계 분석- 대화함축 이해 vs. 화행 이해
The analysis results showed that the correlation between speech acts and conversational implicature was 0.67, indicating a moderately high correlation. As such, speech acts and conversational implicature are both close pragmatic topics in that they are pragmatic interpretations of utterances that take context into account. However, on the other hand, having a moderate correlation implies that there are subtle differences between two linguistic phenomena in their mechanism behind \citep{austin, searle69}. In the case of conversational implicature, all four types have a common underlying principle, i.e. Grice’s cooperative principle \citep{grice}, but in the case of speech acts, each type of speech act, such as directive, expressive, and commissive speech acts, has different felicity conditions, a set of criteria that must be satisfied for a particular speech act to be successfully performed.\footnote{This concept, introduced by J.L. Austin and further developed by John Searle, plays a crucial role in determining whether a speech act is valid and effective within its context.}

% 상관관계 분석- 대화함축 이해 vs. 직접/간접화행 이해
According to the Table \ref{tab:discussion_correlation}, compared to direct speech act, the understanding of indirect speech acts is more correlated with that of implicatures in the sense of implicit comprehension, showing 0.73 of correlation score. In particular, the understanding of indirect speech acts is more correlated with implicit comprehension caused by the maxim of relation.

% LLM의 축어적 의미 의존성 큼
% 대화함축(질의격률 위반 함축 우수)
% 간접화행(정보제공형 지시화행 열등) 
As discussed in section 4.A.b., the relatively low comprehension of directive indirect speech acts by LLMs can be attributed to their strong reliance on the literal meaning of sentences. When the literal meaning is sufficiently interpretable, LLMs are inclined to interpret the utterance as a direct speech act based on literal meaning rather than recognizing its indirectness. However, when the analysis of the literal meaning fails to yield an acceptable interpretation, LLMs attempt to infer deeper, indirect intentions beyond the surface-level and literal meanings. This aligns with the findings of \cite{park1}, who analyzed the capability of LLMs to understand conversational implicature. According to the paper, most LLMs could easily understand implicatures arising from violations of the maxim of quality, that is, when the truth value of an utterance deviates from reality. In other words, when the literal meaning of a sentence differs from the conversational context or world knowledge, the LLMs were able to capture it and analyze the hidden implicatures beneath the surface meaning of the sentence. This suggests that LLMs with large-scaled training data possess exceptional ability in analyzing the literal meaning of sentences. However, this tendency may pose challenges when performing pragmatics tasks, such as indirect speech acts.

\begin{table*}[ht]
\centering
\begin{tabular}{@{}lcccccccc@{}}
\toprule
\textbf{Model} & \textbf{Speech} & \textbf{Direct} & \textbf{Indirect} & \textbf{Impl.} & \textbf{Quantity} & \textbf{Quality} & \textbf{Relation} & \textbf{Manner} \\  
& \textbf{Act} & \textbf{SA} & \textbf{SA} &  & \textbf{Impl.} & \textbf{Impl.} & \textbf{Impl.} & \textbf{Impl.} \\ \midrule
GPT-3.5 & 50.14 & 63.61 & 36.67 & 51.11 & 66.67 & 52.78 & 42.89 & 53.61 \\
GPT-4 & 71.39 & 84.72 & 58.06 & 65 & 83.89 & 82.22 & 70 & 75.28 \\
Claude3-Haiku & 33.75 & 41.94 & 25.83 & 56.57 & 67.78 & 58.89 & 43.33 & 56.67 \\
Claude3-Sonnet & 64.31 & 81.39 & 47.22 & 62.22 & 81.67 & 67.22 & 54.44 & 66.39 \\
Claude3-Opus & 71.94 & 69.17 & 74.72 & 81.11 & 88.89 & 88.89 & 81.11 & 85 \\
Mistral-small & 65.56 & 83.89 & 47.22 & 57.22 & 57.78 & 54.44 & 35 & 51.11 \\
Mistral-medium & 66.67 & 87.5 & 45.83 & 61.11 & 69.24 & 72.22 & 62.22 & 66.25 \\
Mistral-large & 69.58 & 83.61 & 55.56 & 61.11 & 71.11 & 61.11 & 52.22 & 61.39 \\
Llama-3-8B & 49.44 & 68.33 & 30.56 & 54.44 & 68.89 & 44.44 & 45.56 & 53.33 \\
Qwen-14 & 40.14 & 51.67 & 28.61 & 52.22 & 61.67 & 56.11 & 43.33 & 53.33 \\
Solar-10.7B & 65 & 84.44 & 45.56 & 58.33 & 65.56 & 62.22 & 51.11 & 59.31 \\ \midrule
\textbf{Kendall Speech} & 1.00 &  &  & \textbf{0.65} & 0.67 & 0.51 & 0.64 & 0.53 \\  
\textbf{Act} &  &  &  &  &  &  &  &  \\  
\textbf{Kendall Direct} &  & 1.00 &  & 0.31 & \underline{0.31} & 0.11 & 0.27 & 0.16 \\  
\textbf{SA} &  &  &  &  &  &  &  &  \\  
\textbf{Kendall Indirect} &  &  & 1.00 & 0.73 & 0.66 & 0.65 & \underline{0.7} & 0.59 \\  
\textbf{SA} &  &  &  &  &  &  &  &  \\  
\bottomrule
\end{tabular}
\caption{\label{tab:discussion_correlation}Correlation Score between Speech Act and Conversational Implicature in \cite{park2} (SA: Speech Act, Impl.: Implicature)}
\end{table*}

\subsection*{Context-dependent language comprehension of LLMs}

We conducted a comparative analysis of the evaluation results using PUB, a pragmatic benchmark \citep{sravanthi}, to assess the effectiveness of our test set in evaluating the speech act comprehension capabilities of LLMs. One of the most prominent English test sets was selected for comparison due to the relative scarcity of comprehensive test sets available in the Korean language. PUB primarily focuses on implicatures, presuppositions, and deictic expressions in its test sets. Given that speech acts are most closely aligned with conversational implicatures as pragmatic mechanisms, we compared our test’s scores with those of PUB that specifically evaluate the comprehension of implicated meanings in a dialogue. To ensure consistency, GPT-3.5 was selected as the comparison model, as it was evaluated in both our study and the PUB.

The test scores of GPT-3.5 are as follows (Implicature Natural Language Inference test scores are excluded as they do not pertain to conversational implicatures.):

\begin{itemize}[noitemsep] 
\item Direct/Indirect Response Classification(PUB): 80.2
\item Implicature Recovery(PUB): 78.13
\item Response Classification without Implied Meaning(PUB): 58.18
\item Speech Act Comprehension(Ours): 50.14
\end{itemize}

The ‘Speech Act Comprehension’ test score(50.14) is comparable to that of ‘Response Classification with Implied Meaning’ test(58.18). Both tasks involve a brief context and employ a multiple-choice question format aimed at identifying the speaker’s intention. This similarity in structure and objective suggests that these two tasks are closely related in nature. Consequently, this alignment ensures the reliability of the test set as a valid tool for assessing speech intention comprehension. As a test designed solely to determine whether a specific response is direct or indirect, the ‘Direct/Indirect Response Classification’ test is relatively straightforward. It is, therefore, unsurprising that GPT-3.5 achieved its highest score in this task. However, the high score in the ‘Implicature Recovery’ test is noteworthy, given the absence of descriptive context. It is speculated that the context constructed through multi-turn dialogue may have facilitated GPT-3.5’s ability to comprehend the implied meaning of the last utterance.

\section*{Conclusions}

% 요약- 연구목적, 연구방법
In this study, we evaluated whether LLM appropriately understands the speaker’s utterance intention according to a given conversation situation. Through this, we investigated whether LLM, which has recently gained popularity for its high language ability, also has pragmatic ability, which is considered to be the highest level of language ability. To this end, we conducted an evaluation experiment by constructing a total of 240 Korean test items for three types of indirect speech acts.

% 실험 결과 (요약), 실험 분석 결과 (요약)
Our findings revealed that all LLMs yet showed a lower level of speech act comprehension, compared to humans. In both MCQ and OEQ settings, Claude3-Opus surpassed all other models and it was the only model to show relatively higher understanding of indirect speech acts, similar to humans. 
Furthermore, this study pointed out the possibility of overfitting to English data in Llama3 model, since it provided English response though the prompt was set in Korean. In the case of Solar model, it was observed to be predominantly overfitted to MCQ typed leaderboard benchmarks, creating unnecessary four choices even for OEQs, which is somewhat different from how humans solve OEQ-typed problems. In contrast, Claude3-Opus demonstrated high-leveled sophisticated linguistic abilities, providing plausible, correct answers.

% 향후 연구
%1) 실험 데이터 -> 다른 언어로 번역해 적용
While our study provided a comprehensive evaluation of LLMs’ ability to understand speaker’s intention, to compare with the mainstream benchmarks, test sets in high-resource languages like English are needed. This study constructed evaluation data set grounded in linguistic theory, i.e. Searle’s speech act scheme, and laid optimal foundation for evaluating LLM’s ability for understanding utterance intention. Our future research aims to develop data sets in other languages, such as English, Chinese or German, to validate the robustness of the data sets and to compare the speech act comprehension capabilities of LLMs across different languages.

% 2) 실험 방식 -> 가상 대화 수행
Moreover, we will explore alternative experimental settings by having LLMs to engage in simulated conversations with humans, to evaluate whether they can accurately understand utterance intentions and generate appropriate responses. This study assessed LLMs by having them either to select (MCQ) or explain (OEQ) the intention behind a given utterance considering the given conversational context. However, our future study will focus on evaluating LLMs in natural, conversational settings. Dialogue is an authentic communication format of human language use.
LLMs’ ability could be examined by following two settings: having them to generate next utterance considering the given multi-turn conversation as input prompt (‘Multi-turn Prompting’) or having them to exchange one utterance at a time to evaluate whether the LLM accurately understands the intent behind each utterance and generates a natural response (‘Single-turn Prompting’).

\bibliography{sample}

\end{document}